\documentclass{article}

\usepackage[final]{neurips_2019}

\usepackage{times}
\usepackage{latexsym}
\usepackage{graphicx}
\usepackage{amsmath}
\usepackage{pgfplots}
\usepackage{natbib}
\pgfplotsset{width=0.7\textwidth}

\usepackage{url}

\title{Gender Prediction from Tweets: Improving Neural Representations with Hand-Crafted Features}

\author{%
  Erhan Sezerer\\
  Dept. of Computer Engineering\\
  \.{I}zmir Institute of Technology\\
  \.{I}zmir, Turkey \\
  \texttt{erhansezerer@iyte.edu.tr} \\
  \And
  Ozan Polatbilek \\
  Dept. of Computer Engineering\\
  \.{I}zmir, Turkey \\
  \texttt{ozanpolatbilek@iyte.edu.tr} \\
  \AND
  Selma Tekir \\
  Dept. of Computer Engineering\\
  \.{I}zmir, Turkey \\
  \texttt{selmatekir@iyte.edu.tr} \\
}

\date{}
\pgfplotsset{compat=1.14}
\begin{document}

\maketitle

\begin{abstract}
  Author profiling is the characterization of an author through some key attributes such as gender, age, and language. In this paper, a RNN model with Attention (RNNwA) is proposed to predict the gender of a twitter user using their tweets. Both word level and tweet level attentions are utilized to learn 'where to look'. This model\footnote{https://github.com/Darg-Iztech/gender-prediction-from-tweets} is improved by concatenating LSA-reduced n-gram features with the learned neural representation of a user. Both models are tested on three languages: English, Spanish, Arabic. The improved version of the proposed model (RNNwA + n-gram) achieves state-of-the-art performance on English and has  competitive results on Spanish and Arabic. 
\end{abstract}

\section{Introduction}

Author profiling is the characterization of an author through some key attributes such as gender, age, and language. It's an indispensable task especially in security, forensics, and marketing. Recently, social media has become a great data source for the potential learning approaches. Furthermore, gender prediction has been a popular profiling task.

The traditional approach to gender prediction problem is extracting a useful set of hand-crafted features and then feeding them into a standard classification algorithm. In their study, \cite{handcraft-2} work with the style-based features of message length, stop word usage, frequency of smiley etc. and use different classifiers such as k-nearest neighbor, naive bayes, covering rules, and backpropagation to predict gender on chat messages. Similarly, \cite{email-hancraft} select some hand-crafted features and feed them into various classifiers.

Most of the work on gender prediction rely on n-gram features \citep{miller-ngram}. \cite{daneshvar2018} give Latent Semantic Analysis (LSA)-reduced forms of word and character n-grams into Support Vector Machine (SVM) and achieve state-of-the-art performance. Apart from exploiting n-gram frequencies, there are studies \citep{language-independent}, \citep{Language-Independent-2013}, \citep{bleaching-text} to extract cross-lingual features to determine gender from tweets. Some other work \citep{language-independent}, \citep{metadata-used} exploit user metadata besides using just tweets.

Recently, neural network-based models have been proposed to solve this problem. Rather than explicitly extracting features, the aim is to develop an architecture that implicitly learns. In author profiling, both style and content-based features were proved useful \citep{argamon2009} and neural networks are able to capture both syntactic and semantic regularities.  In general, syntactic information is drawn from the local context. On the other hand, semantic information is often captured with larger window sizes. Thus, CNNs are preferred to obtain style-based features while RNNs are the methods of choice for addressing content-based features \citep{goldberg2017}. In literature, CNN \citep{selfcite} or RNN \citep{takahashi2O18}, \citep{Kodiyan}, \citep{selfcite_SIU} is used on this task. \cite{takahashi2O18} obtain state-of-the-art performance among neural methods by proposing a model architecture where they process text through RNN with GRU cells. Also, the presence of an attention layer is shown to boost the performance of neural methods \citep{takahashi2O18}, \citep{selfcite}.

In this work, we propose a model that relies on RNN with attention mechanism (RNNwA). A bidirectional RNN with attention mechanism both on word level and tweet level is trained with word embeddings. The final representation of the user is fed to a fully connected layer for prediction. Since combining some hand-crafted features with a learned linear layer has shown to perform well in complex tasks like Semantic Role Labeling (SRL) \citep{collobert2008}, an improved version of the model (RNNwA + n-gram) is also tested with hand-crafted features. In the improved version, LSA-reduced n-gram features are concatenated with the neural representation of the user. Then the result is fed into a fully-connected layer to make prediction. Models are tested in three languages; English, Spanish, and Arabic, and the improved version achieves state-of-the-art accuracy on English, and competitive results on Spanish and Arabic corpus.

There are many datasets created for this task \citep{2018overview}, \citep{selfcite_ACL}. In this work, we have used the dataset and benchmarks provided by the PAN 2018 shared task on author profiling \citep{2018overview}. As the dataset contains a constant number of $100$ tweets per user, accuracy tests are performed both on user and tweet level (tweet-level predictions are made by removing the user-level attention). Tweet-level accuracy tests show interesting results during hyperparameter optimization. When the tweet-level predictions are averaged to produce user-level predictions, it is seen that the hyperparameters that gave the best results in terms of tweet-level accuracy, performs worse in user-level accuracy. The better user-level models, with different hyperparameters, that gave the highest user-level accuracy are observed to slightly overfit on tweet-level. It leads us to believe that the overfitting in the tweet-level predictions in best user-level models acts similar to an attention mechanism by over-emphasizing some distinctive tweets and ignoring the rest.

\section{Model architecture}
In author profiling, both style-based and content-based features must be addressed \citep{argamon2009}. An appropriate baseline for this task is a CNN-based model that is able to capture style-based information \citep{selfcite}. The proposed RNN-based model relies on extracting content-based features. In addition, in order to improve its accuracy, the proposed model is combined with some hand-crafted features. For all of the models, Adam optimizer \citep{adam} is used with cross-entropy loss along with the L2 regularization to prevent from overfitting.

\begin{figure}[t]
    \centering
 	\includegraphics[width=1\textwidth]{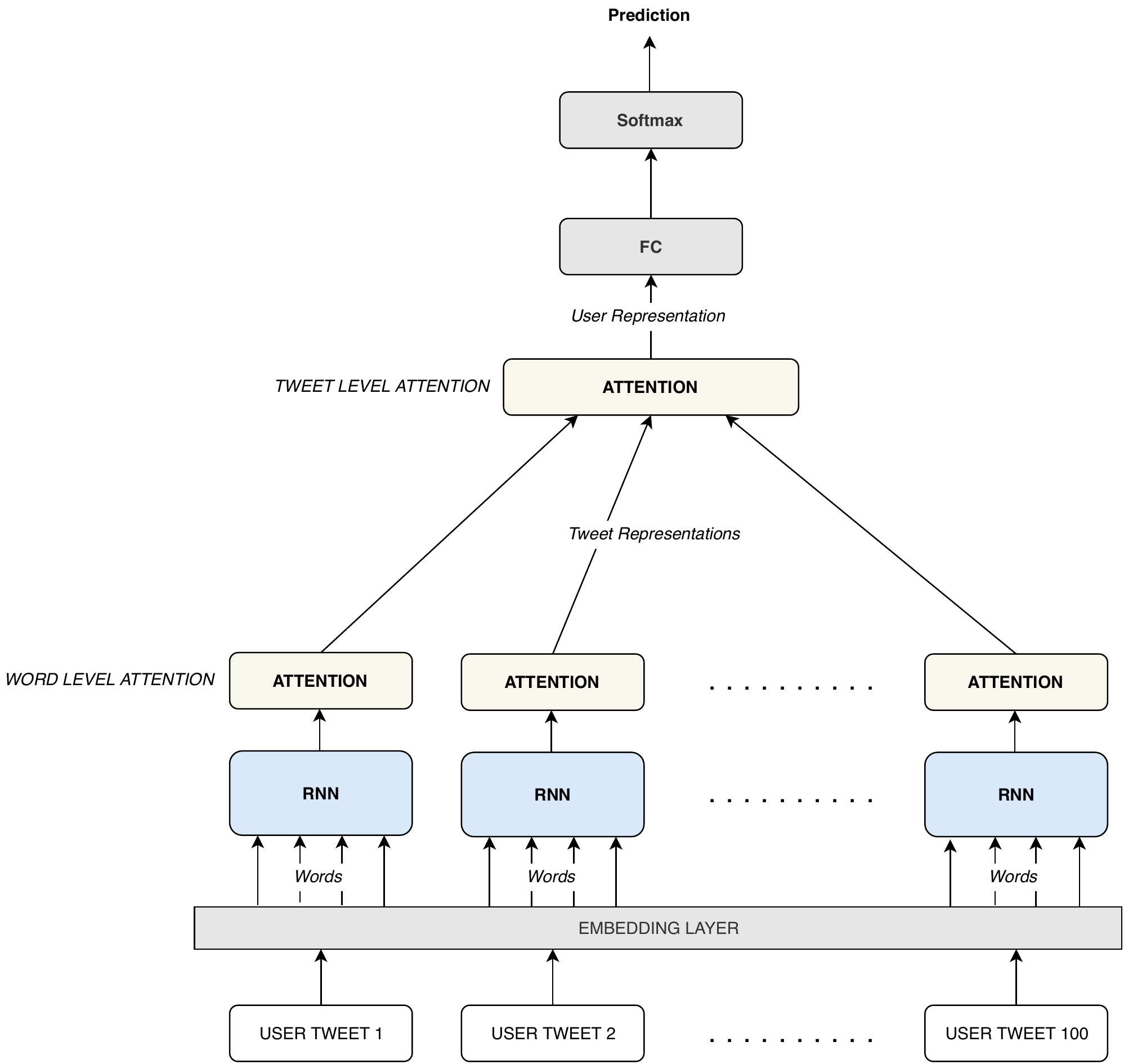}
    		\caption{Proposed model.}
     		\label{fig:model}
\end{figure}

\subsection{Baseline CNN model}

CNN model (denoted CNNwA on results) is based on \cite{selfcite} where each character in the tweet is represented with a character embedding of size $25$, which is trained along the neural network. All characters are lower-cased. Non-alphabetical characters such as punctuation are kept with a view to capturing some information on the profile of the user since they are heavily used in twitter as emoticons.

Filters of size $3\times3$, $6\times6$ and $9\times9$ are used for each language, and the number of filters is determined by performing grid search on validation set. Among the tested range ($50$-$125$ with intervals of $25$), the number of filters that gives the best accuracy is $100$ (per each filter), for all languages.

\begin{figure}[t]
    \centering
 	\includegraphics[width=1\textwidth]{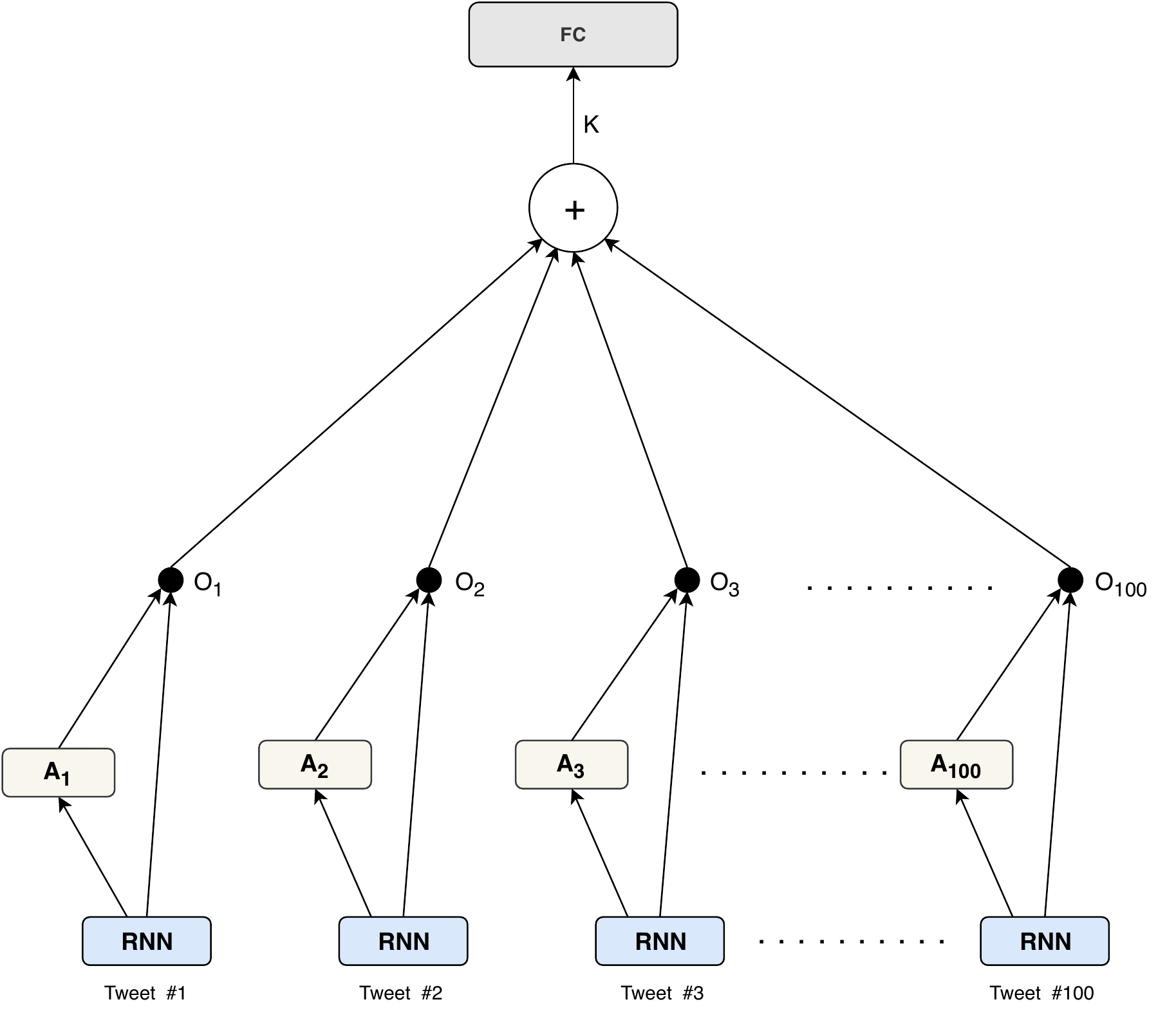}
    		\caption{Tweet-level Attention Layer in Detail.}
     		\label{fig:attention}
\end{figure}

\subsection{RNN Model}

Since the dataset is not big enough to train word embeddings, Glove word embeddings \citep{glove} of size $200$ are used in the proposed RNN Model (denoted RNNwA on results) due to their success at various NLP tasks and their multi-linguality: They encompass all the languages in the test set. In addition, the Glove embeddings are also trained on Twitter data which make them reflect the nature of the dataset better than other alternatives.

A bidirectional RNN with GRU \citep{GRU} cells are used in this model where the number of cells is a hyperparameter. Among the tested range ($50$-$150$ with intervals of $25$), best accuracy on validation set is obtained by $150$ cells in English and $100$ cells in Spanish and Arabic. An attention mechanism is used on word-level in addition to tweet-level to capture the important parts of each tweet as shown in Figure \ref{fig:model}.

A feature vector for each tweet is created by feeding tweets to RNN separately. In order to discriminate tweets with respect to their information carrying capacity on its author's gender, Bahdanau attention mechanism \citep{attention} is used to combine the tweets rather than concatenating them before feeding to the network or averaging their predictions later. Figure \ref{fig:attention} shows the tweet-level attention layer in detail which is calculated by the following formulas:

\begin{equation}
\nonumber
\begin{split}
A_i &= tanh(\boldsymbol{W_\alpha} t_i+b) \\
v_i &= \frac{exp(A_iw_i)}{\sum_{j}^{}{exp(A_jw_j)}} \\
o_i &= v_i t_i \\
K &= \sum_{i}^{}{o_i} \\
\end{split}
\end{equation}

where $\boldsymbol{W_\alpha}$ is a learnable weight matrix that is used to multiply each output of the RNN, $t_i$ is the feature vector of $i$th tweet, $b$ is a learnable bias vector, $w_i$ is a learnable attention weight, $A_i$ is the attention context vector, $v_i$ is the attention value for $i$th tweet, $o_i$ is attention output vector for the corresponding tweet, $K$ is the output vector for user. Matrix $\boldsymbol{W_\alpha}$ and vectors $w_i$ and $b$ are learned parameters.

Attention layer outputs a single feature vector that corresponds to a user, which is then fed to a fully-connected layer to lower the dimension to the number of classes.

There are two different attention layers on the model. One is a word level attention where it amplifies the signal coming from important words, the other one is on tweet level where it combines the signals coming from each tweet and creates the final representation of a user.

\subsection{RNN with N-gram Model}
For this model (denoted RNNwA + n-gram on results), n-gram features are collected with the same method described in \cite{daneshvar2018}. At the beginning, word level and character level n-gram features are obtained and concatenated. Then they are normalized with tf-idf transformation. For reducing the number of features and sparsity in n-gram vectors, tuples that have frequency less than $2$ are ignored. For character level n-gram $N$ is selected as $3,4$, and $5$ and for word level n-gram, $N$ is $1,2$ for Spanish and Arabic; $1,2,3$ for English. The dimension of the vector is reduced by LSA to $300$. Then the vector is concatenated with neural representation which is produced right after tweet level attention in RNNwA model. The resultant representation is fed to a fully- connected layer that produces predictions. 

\subsection{Dataset}

Models are tested on the PAN 2018 author profiling dataset \citep{2018overview}, which provides tweets in three languages: English, Spanish and Arabic with training/test datasets of sizes ($3000$ users, $1900$ users), ($3000$ users, $2200$ users), and ($1500$ users, $1000$ users) respectively, where each user has $100$ tweets. Each training set is further partitioned randomly into training and validation sets with the ratio ($0.8$, $0.2$) respectively for hyper-parameter optimization.

\section{Results}
In order to measure the effectiveness of the attention mechanism, in addition to the CNN baseline model (CNNwA) and RNNwA, two new models (denoted as CNN and RNN) are created by removing the tweet level attention layer (word level attention stays the same) and generating a prediction for each tweet then just simply taking an average to give a user level prediction. Tweet level accuracies for these models are shown in Table \ref{tab:resultstweet}.

\begin{table}[t]
	\centering
	\caption{Tweet Level Accuracy of the CNN and RNN  Models without Attention.}\label{tab:resultstweet}
	\begin{tabular}{c|ccc}
		\hline
		Model & English & Spanish & Arabic\\
		\hline
		RNN   & 62.600 & 62.163 & 62.170 \\
		CNN   & 59.675 & 59.700 & 59.267 \\
		\hline
	\end{tabular}
\end{table}

\begin{table}[t]
	\centering	
	\caption{User Level Accuracy of the Proposed Model (RNNwA) along with the Baselines.}\label{tab:resultsuser}
	\begin{tabular}{c|ccc}
		\hline
		Model & English & Spanish & Arabic\\
		\hline
		CNN   & 74.947 & 71.772 & 72.100 \\
		CNNwA & 78.474 & 75.000 & 71.800\\
		RNN   & 79.316 & 74.091 & 77.100 \\
		RNNwA & 81.789 & 78.227 & 78.500 \\
		\hline
	\end{tabular}
\end{table}

In Table \ref{tab:resultsuser}, user level accuracy results for the proposed model (RNNwA) along with the baseline models are given. As can be seen in the results, tweet level attention mechanism increases the score of all baseline models with the only exception of the CNNwA model in Arabic.

Also, compared to the best neural model \citep{takahashi2O18} where max pooling is used instead of an attention mechanism on the outputs of RNN, the proposed model (RNNwA) gives better results in terms of accuracy on English and Arabic datasets, and produces similar accuracy levels on Spanish dataset (Table \ref{tab:comparison}). These results show that an attention layer is able to learn "where/how to look" for features that are helpful in identifying the gender of a user.

On the other hand, the improved model (RNNwA + n-gram), where neural and hand-crafted features are concatenated, increases the accuracy of the proposed model by approximately $0,5$\% on English and approximately $2$\% in Spanish and Arabic. This also supports our intuition that the performance of neural models can be improved by hand-crafted features, which is based on the study of \cite{collobert2008}. As can be seen in Table \ref{tab:comparison}, the improved model outperforms the state-of-the-art method of \cite{daneshvar2018} in English and produces competitive results in Spanish and Arabic.

\begin{figure}[hb]
    \centering
    \begin{tikzpicture}
\begin{axis}[
    title={},
    xlabel={Tweet Level Accuracy},
    ylabel={User Level Accuracy},
    xmin=57, xmax=64,
    ymin=67, ymax=82,
    xtick={58,59,60,61,62,63},
    ytick={69,71,73,75,77,79,81},
    legend pos=north east,
    ymajorgrids=true,
    xmajorgrids=true,
    grid style=dashed,
    legend entries={
            English,
            Spanish,
            Arabic,%
        },
]
\addplot+[
    scatter, only marks,
    scatter/@pre marker code/.append code={%
        \let\pgfplotspointmeta=\partition
    },%
    scatter/classes*={
        dummy_en={mark=square,black},
        dummy_es={mark=triangle,black},
        dummy_ar={mark=o,black},
        ENGLISH={mark=square,red},%
        SPANISH={mark=triangle,red},%
        ARABIC={mark=o,red},%
        ENGLISHTWEET={mark=square,blue},%
        SPANISHTWEET={mark=triangle,blue},%
        ARABICTWEET={mark=o,blue}},%
    %
    visualization depends on={value \thisrow{partition} \as \partition},
]
    table[meta=partition]{
        x         y      partition 
        62.6    75.684     ENGLISHTWEET
        62.592  76.263     ENGLISHTWEET
        62.583  75.263     ENGLISHTWEET
        59.146  79.315     ENGLISH
        59.554  78.842     ENGLISH
        59.333  78.789     ENGLISH
        62.163  69.227     SPANISHTWEET
        62.157  69.272     SPANISHTWEET
        62.148  69.318     SPANISHTWEET
        58.928  74.091     SPANISH
        58.830  74.045     SPANISH
        61.480  72.909     SPANISH
        62.170  72.9       ARABICTWEET
        62.157  72.9       ARABICTWEET
        62.147  73.3       ARABICTWEET
        57.786  77.1       ARABIC
        58.143  76.3       ARABIC
        57.964  76.4       ARABIC
    };
\end{axis}
\end{tikzpicture}
    \caption{Comparison of Tweet-Level and User-level accuracy of RNN Model. Best three user-level models (colored in red) and best three tweet-level models (colored in blue) are selected for each language.}
    \label{fig:scatterplot}
\end{figure}
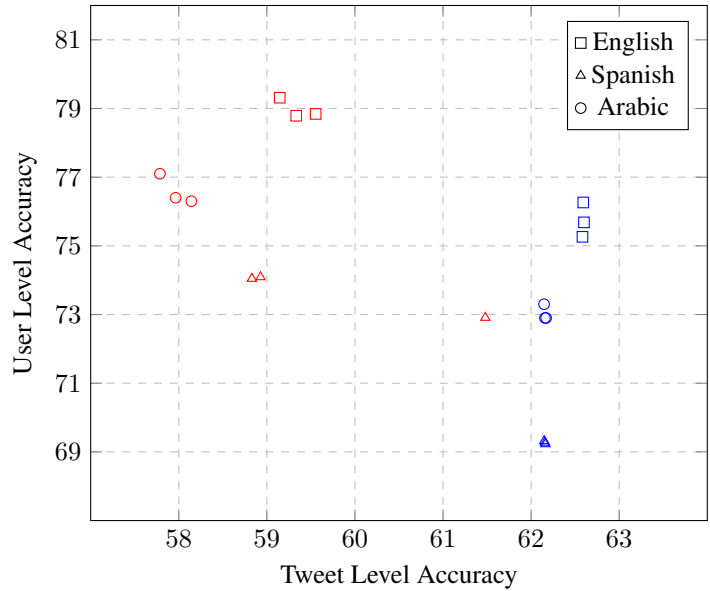

There is an interesting observation concerning the models without tweet level attention (RNN and CNN) in hyper-parameter optimization. During the hyperparameter optimization of the models RNN and CNN, we saved both the models that gave the best tweet-level accuracy and the models that gave the best user-level accuracy. The expectation is to see that the best setup on tweet-level also gives the best performance in user-level, but the outcome is the opposite: Best setups on tweet-level always fall behind best user-level setups. Performance differences between various setups can be seen in Figure \ref{fig:scatterplot} where accuracies of the best three models in terms of tweet-level and best three models in terms of user-level are shown for all languages. It can be observed that the best tweet-level setups are almost $4\%$ worse in terms of user-level accuracy. Deeper investigation shows that the best user-level models exhibit slight overfitting on tweet-level, in training. Although overfitting normally leads to poor generalization, in this case we believe that this overfitting acts similar to an attention mechanism by over-emphasizing some important tweets and ignoring uninformative ones in the process. Even though this leads to poor tweet-level accuracy, it improves the user-level accuracy of the models as it can be seen from the Figure \ref{fig:scatterplot}.

\begin{table}[t]
	\centering
	\caption{Accuracy on PAN 2018 test set.}\label{tab:comparison}
	\begin{tabular}{c|ccc}
		\hline
		Model & English & Spanish & Arabic\\
		\hline
		\cite{daneshvar2018}\footnotemark[1] &  81.52  & \textbf{82.00} & \textbf{80.90} \\
		\cite{takahashi2O18}\footnotemark[2] & 79.68 & 78.64 & 77.10 \\
		Proposed Model (RNNwA) & 81.79 & 78.23 & 78.50 \\
		Improved Model (RNNwA + n-gram) & \textbf{82.31} & 80.22 & 80.50 \\
		\hline
	\end{tabular}
\end{table}
\footnotetext[1]{In their paper, authors report a result of 82.21 in English but we couldn't verify their accuracy in our repetitions by using their software and the same dataset.}
\footnotetext[2]{Since their software is not provided, we directly take the accuracy values from their paper.}

\vspace{0,5cm}

\vspace{0,5cm}

\section{Conclusion}

In this work, a neural network-based model namely RNN with attention (RNNwA) is proposed on the task of gender prediction from tweets. The proposed model is further improved by hand-crafted features which are  obtained by LSA-reduced n-grams and concatenated with the neural representation from RNNwA. User representations that is the result of this model is then fed to a fully-connected layer to make prediction. This improved model achieved state-of-the-art accuracy on English and has a competitive performance on Spanish and Arabic.

We also would like to kindly remind our readers that although the model is self-learning, there might still exist a gender bias in the evaluation of the model due to the data itself. Since the model learns to predict the gender directly from tweets of the twitter users, any bias the twitter users have might be reflected in the model predictions.

\section*{Acknowledgments}
We would like to thank Computer Vision Research Group from Izmir Institute of Technology for providing us the hardware for performing the tests in this research.

The Titan V used for this research was donated by the NVIDIA Corporation.

\bibliography{arxiv}

\end{document}